\documentclass[journal,10pt]{IEEEtran}
\usepackage{scrpage}
\usepackage{amsmath}                        
\usepackage{booktabs}                       
\usepackage{epigraph}
\usepackage[all]{xy}
\usepackage[hang]{subfigure}                
\usepackage{url}                            
\usepackage{amssymb, cite, amsmath, subfigure, stfloats, amssymb, amsfonts, rotating, acronym}
\usepackage[para]{threeparttable}
\usepackage{multirow}
\usepackage{array}
\usepackage{threeparttable}
\usepackage{tabularx}
\usepackage{graphicx}
\usepackage{epstopdf}
\usepackage[percent]{overpic}
\usepackage[T1]{fontenc}
\usepackage[utf8]{inputenc}
\usepackage{authblk}
\usepackage[]{algorithm2e}

\newcommand{\E}{\mathbb{E}}
\newcommand{\OR}{\text{OR}}
\newcommand{\AND}{\text{AND}}
\newcommand{\XNOR}{\text{XNOR}}

\PassOptionsToPackage{bookmarks={false}}{hyperref}
\begin{document}

\title{VLSI Implementation of Deep Neural Network Using Integral Stochastic Computing}
\author{Arash~Ardakani, \emph {Student Member, IEEE}, Fran{\c{c}}ois Leduc{-}Primeau, Naoya Onizawa, \emph {Member, IEEE}, \\ Takahiro Hanyu, \emph{Senior Member, IEEE} and Warren~J.~Gross, \emph{Senior Member, IEEE }
\thanks{A preliminary version of this paper was published in \cite{mypaper}.}}

\markboth{}{}

\maketitle

\begin{abstract}
The hardware implementation of deep neural networks (DNNs) has recently received tremendous attention: many applications in fact require high-speed operations that suit a hardware implementation. However, numerous elements and complex interconnections are usually required, leading to a large area occupation and copious power consumption. Stochastic computing has shown promising results for low-power area-efficient hardware implementations, even though existing stochastic algorithms require long streams that cause long latencies. In this paper, we propose an integer form of stochastic computation and introduce some elementary circuits. We then propose an efficient implementation of a DNN based on integral stochastic computing. The proposed architecture has been implemented on a Virtex7 FPGA, resulting in 45\% and 62\% average reductions in area and latency compared to the best reported architecture in literature. We also synthesize the circuits in a 65~nm CMOS technology and we show that the proposed integral stochastic architecture results in up to 21\% reduction in energy consumption compared to the binary radix implementation at the same misclassification rate. Due to fault-tolerant nature of stochastic architectures, we also consider a \emph{quasi-synchronous} implementation which yields 33\% reduction in energy consumption w.r.t. the binary radix implementation without any compromise on performance.
\end{abstract}


\section{Introduction}\label{introduction}
Recently, the implementation of biologically-inspired artificial neural networks such as the Restricted Boltzmann Machine (RBM) has 
aroused great interest due to their high performance in approximating complicated functions. 
A variety of applications can benefit from them, in particular machine learning algorithms. They can be split in two phases, which are referred to as \emph{learning} and \emph{inference} phases \cite{ref9}. The learning engine finds a proper configuration to map learning input data into their desired outputs, while the inference engine uses the extracted configuration to compute outputs for new data.\par

\begin{figure}[!t]
\centering
\includegraphics[scale = 1]{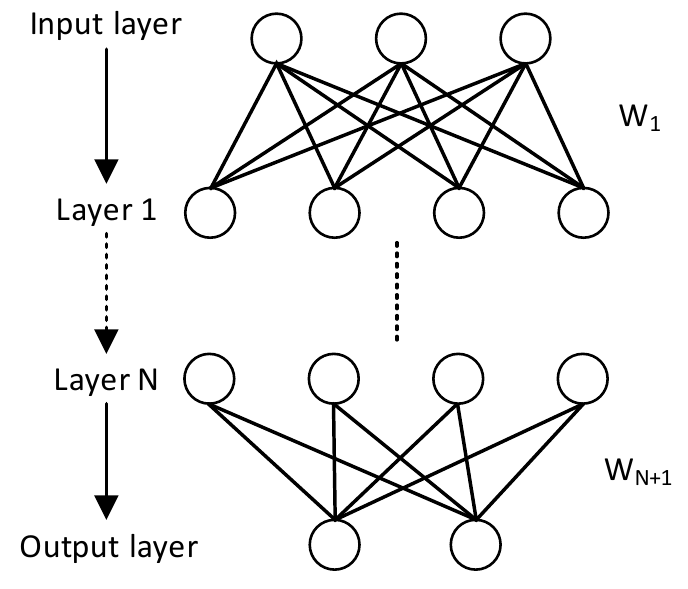}
\caption{A N-layer DBN where $W$ and $N$ denote the weights of each layer and number of layers respectively.}
\label{DBN}
\end{figure}

Deep neural networks, especially Deep Belief Networks (DBN), have shown state-of-the-art results on various computer vision and recognition tasks \cite{ref10,ref1,ref2,ref11,ref12,ref13}. DBN can be formed by stacking RBMs on top of each other to construct a deep network, as shown in Fig.~\ref{DBN} \cite{ref1}. RBMs used in DBN are pre-trained using Gradient-based Contrastive Divergence (GCD) algorithms, followed by gradient descent and backpropagation algorithms for classification and fine-tuning the results \cite{ref1,ref2}. \par
In the past few years, general purpose processors have been mainly used for software realization of both training and inference engines of DBN. However, large power consumption and high resource utilization have pushed researchers to explore ASIC and FPGA implementations of neural networks. Rapid expansion of devices and sensors connected to the internet of things (IoT) allows to perform the training procedure once on cloud servers equipped with Graphics Processing Unit (GPU), and extract weights for inference engine usage through the IoT platforms. The inference engine can then be implemented using ASIC or FPGA platforms.\par 
\par
DBNs are constructed of multiple layers of RBMs and a classification layer at the end. The main computation kernel consists of hundreds of vector-matrix multiplications followed by non-linear functions in each layer. Since multiplications are costly to implement in hardware, existing parallel or semi-parallel VLSI implementations of such a network suffer from high silicon area and power consumption \cite{ref24}. The nonlinearity function is also implemented using Look-Up Tables (LUTs), requiring large memories. Moreover, hardware implementation of this network results in large silicon area: 
this is caused by the connections between layers, that lead to severe routing congestion.
Therefore, an efficient VLSI implementation of DBN is still an open problem.\par

Recently, Stochastic Computing (SC) has shown promising results for ultra low-cost and fault-tolerant hardware implementation of various systems \cite{ref14,ref15,ref16,ref31,ref32,ref33,ref34,ref35,ref36,ref37}. Using SC, many computational units have simple implementation. For instance, using unipolar SC, the multiplication and addition are implemented using an $\textsc{and}$ gate and a multiplexer, respectively \cite{ref3,ref4}. However, the multiplexer-based adder introduces a scaling factor that can cause a precision loss \cite{ref5}, resulting in the failure of SC for deep neural networks, which require many additions. An $\textsc{or}$ gate can provide a good approximation to addition if its input values are small \cite{ref4}. However, using $\textsc{or}$ gates to perform addition in DBNs results in a huge misclassification error compared to its fixed-point hardware implementation. Therefore, an efficient stochastic implementation that maintains the performance of DBN is still missing.\par

In this paper, an integral stochastic computation is introduced to solve the precision loss issue of conventional scaled-adder, while also reducing the latency compared to conventional binary stochastic computation. It is also worth mentioning that the proposed technique results in lower latency compared to conventional binary stochastic computation. A novel Finite State Machine (FSM)-based $\tanh$ function is then proposed as the nonlinearity function used in DBN. Finally, an efficient stochastic implementation of DBN based on the aforementioned techniques with an acceptable misclassification error is proposed, resulting in 45\% smaller area on average compared to the state-of-the-art stochastic architecture.\par

A nanoscale memory-resistor (memristor) device is a non-volatile digital memory, which consumes substantially less energy compared to CMOS and can be scaled to sizes below 10~nm \cite{ref25}. A challenging problem with memristor devices is the presence of significant random variations. A promising approach for dealing with the non-determinism of memristors is to design SC systems that are fault-tolerant \cite{ref25}. In this paper, we show that the proposed architectures can tolerate a fault rate of up to 16\% when timing violations are allowed to occur, making them suitable for memristor devices.\par

The manuscript can be divided in two major parts: the proposed algorithms and their hardware implementation results. In the first part, we analyze elementary computational units. Also, some simulation results and examples are provided to shed light on the proposed algorithm in comparison with the existing methods. In the second part, design aspects of a deep neural network based on the proposed method are studied and some implementation results under different conditions are provided.\par 

The rest of this paper is organized as follows. Section \ref{sec1} provides a review of SC and its computational elements. Section \ref{sec2} introduces the proposed integral stochastic computation and operations in this domain. Section \ref{sec3} describes the integral stochastic implementation of DBN. Implementation results of the proposed architecture is provided in Section \ref{sec4}. In this section, the performance of the stochastic implementation is studied when the circuit is affected by timing violations. Note that accepting occasional timing violations allows to reduce the supply voltage, which can improve the energy efficiency of the system. In Section \ref{sec5}, we conclude the paper and discuss future research.
\begin{figure}[!t]
    \centering
    \subfigure[]{
        \includegraphics[]{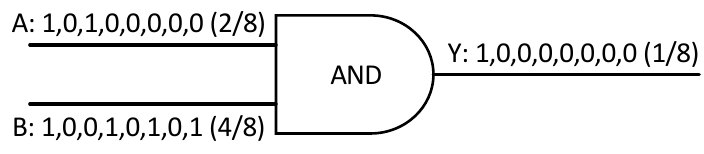}
        \label{AND}
    }
    \subfigure[]{
        \includegraphics[]{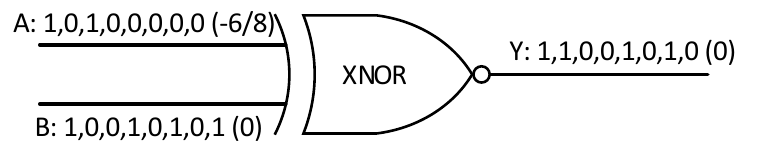}
        \label{XNOR}
    }
    \caption{Stochastic multiplications using (a) \textsc{and} gate in unipolar format and (b) \textsc{xnor} gate in bipolar format}
    \label{mult}
\end{figure}

\section{Stochastic Computing and Its Computational Elements}\label{sec1}
In stochastic computation, numbers are represented as sequences of random bits. The information content of the sequence does not depend on the particular value of each bit, but rather on their statistics. Let us denote by $X \in \{0, 1\}$ a bit in the random sequence. To represent a real number $x \in [0, 1]$, we simply generate the sequence such that:
\begin{equation}
\E[X]= x,
\end{equation}where $\E[X]$ denotes the expected value of the random variable $X$. This is known as the \emph{unipolar} format. The \emph{bipolar} format is another commonly used format where $x \in [-1, 1]$ is represented by setting:
\begin{equation}
\E[X] = (x+1)/2.
\end{equation}
Note that any real number can be represented in one of these two formats by scaling it down to fit within the appropriate interval. In this paper, we use upper case letters to represent elements of a stochastic stream, while lower case letters represent the real value associated with that stream. It is also worth mentioning that a stochastic stream of a real value $x$ is usually generated by a linear feedback shift register (LFSR) and a comparator. This unit is hereafter referred to as binary to stochastic convertor (B2S) \cite{ref27}.\par


\subsection{Multiplication In SC}\label{subsec1}
Multiplication of two stochastic streams is performed using \textsc{and} and \textsc{xnor} gates in unipolar and bipolar encoding formats, respectively, as illustrated in Fig.~\ref{AND} and \ref{XNOR}. In unipolar format, the multiplication of two input stochastic streams of $A$ and $B$ is computed as:
\begin{equation}
Y = \AND\left(A,B\right) = A \cdot B,
\end{equation}
where "$\cdots$" denotes bit-wise \textsc{and} and if the input sequences are independent, we have:
\begin{equation}
y = \E[Y] = a \times b.
\end{equation}
Multiplications in bipolar format can be performed as:
\begin{equation}
Y = \XNOR\left(A,B\right) = \OR\left(A\cdot B,\left(1-A\right)\cdot\left(1-B\right)\right),
\end{equation}
\begin{equation}
\E[Y] = \E[A\cdot B] + \E[(1-A)\cdot(1-B)].
\end{equation}
If the input streams are independent,
\begin{equation}
\E[Y] = \E[A]\times \E[B] + \E[1-A]\times \E[1-B].
\end{equation}
By simplifying the above equation, we have:
\begin{equation}
y = 2\E[Y]-1 = \left(2\E[A]-1\right)\times\left(2\E[B]-1\right).
\end{equation}

\begin{figure}[!t]
    \centering
    \subfigure[]{
        \includegraphics[scale = 0.8]{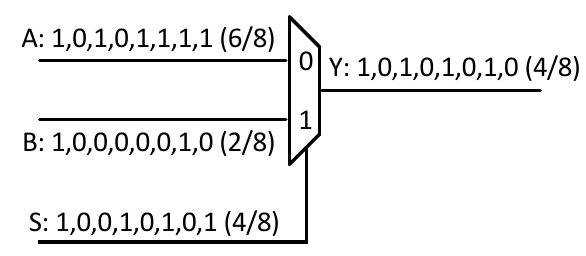}
        \label{MUX}
    }
    \subfigure[]{
        \includegraphics[scale = 0.8]{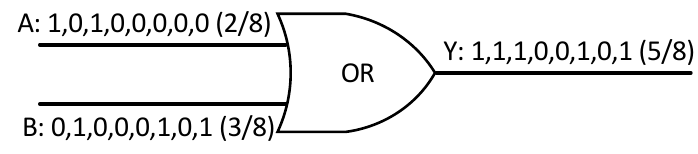}
        \label{OR}
    }
    \caption{Stochastic additions using (a) MUX and (b) \textsc{or} gate}
    \label{adder}
\end{figure}

\subsection{Addition In SC}\label{subsec2}
Additions in SC are usually performed by using either scaled adders or \textsc{or} gates \cite{ref3,ref4}. The scaled adder uses a multiplexer (MUX) to perform addition. The output of a MUX $Y$ is given by
\begin{equation}
Y = A\cdot S + B\cdot\left(1-S\right).
\label{add}
\end{equation}
As a result, the expected value of $Y$ would be $(\E[A]+\E[B])/2$ when the select signal $S$ is a stochastic stream with probability of 0.5, as illustrated in Fig.~\ref{MUX}. This 2-input scaled adder ensures that its output is in the legitimate range of each encoding format by scaling it down by factor of 2. Therefore, $L$-input addition can be performed by using a tree of multiple 2-input MUXs. In general, the result of an $L$-input scaled adder is scaled down $L$ times, which can decrease the precision of the stream. To achieve the desired accuracy, longer bit-streams must be used, resulting in larger latency.\par

\textsc{or} gates can also be used as approximate adders as shown in Fig.~\ref{OR}. The output $Y$ of an $\textsc{or}$ gate with inputs $A$, $B$ can be expressed as
\begin{equation}
Y = A + B - A\cdot B.
\label{add1}
\end{equation}
\textsc{or} gates function as adders only if $\E[AB]$ is close to 0. Therefore, the inputs should first be scaled down to ensure that the aforementioned conditions are met. This type of adder still requires long bit-streams to overcome a precision loss incurred by the scaling factor.
\par
To overcome this precision loss, which could potentially lead to inaccurate results, the Accumulative Parallel Counter (APC) is proposed in \cite{ref5}. The APC takes $N$ parallel bits as inputs and adds them to a counter in each clock cycle of the system. Therefore, this adder results in lower latency due to its small variance of the sum. It is also worth mentioning that this adder converts the stochastic stream to binary form \cite{ref5}. Therefore, this adder is restricted to cases where additions are performed to obtain the final result, or requiring an intermediate result in binary format.\par

\begin{figure}[!t]
    \centering
    \subfigure[]{
        \includegraphics[scale = 0.5]{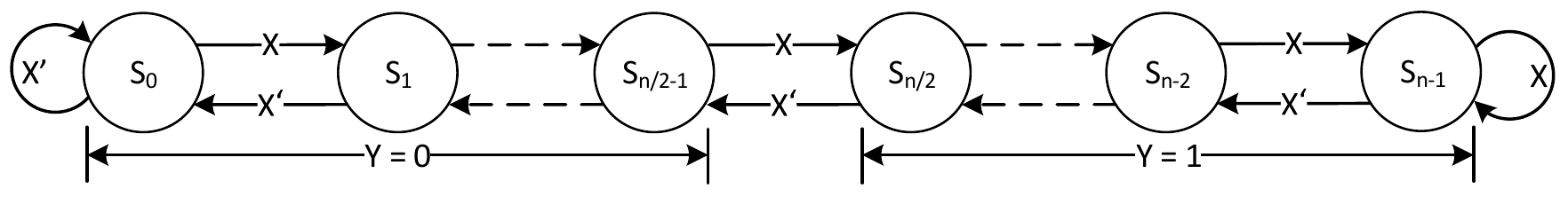}
        \label{TANH}
    }
    \subfigure[]{
        \includegraphics[scale = 0.5]{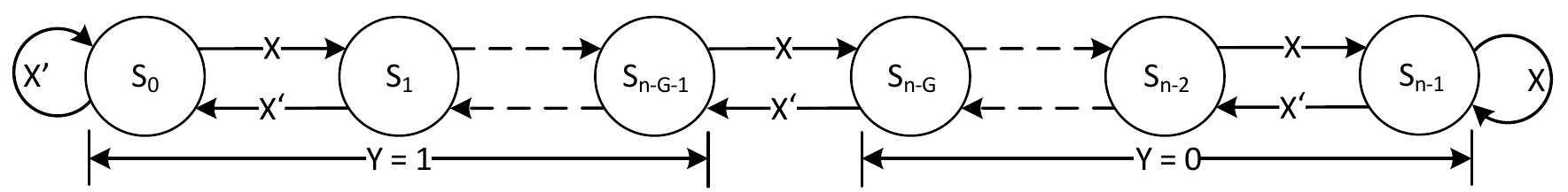}
        \label{EXP}
    }
    \caption{State transition diagram of the FSM implementing (a) $\tanh$ and (b) exponentiation functions}
    \label{fsm}
\end{figure}

\subsection{FSM-Based Functions In SC}\label{subsec3}
Hyperbolic tangent and exponentiation functions are 
computations required by many applications.
These functions are implemented in the stochastic domain by using a FSM \cite{ref6}. Fig.~\ref{TANH} and \ref{EXP} show the state transition diagram of the FSM implementing $\tanh$ and exponentiation functions. The FSM is constructed such that
\begin{equation}
\tanh\left(\dfrac{nx}{2}\right) \approx \E[\text{Stanh}\left(n,X\right)],
\label{tanh}
\end{equation}
\begin{equation}
\exp\left(-2Gx\right) \approx \E[\text{Sexp}\left(n,G,X\right)] : x > 0.
\label{exp}
\end{equation}
where $n$ denotes the number of states in the FSM, $G$ the linear gain of the exponentiation function and $Y$ the stochastic output sequence. 
Let us define as Stanh and Sexp the approximated functions of $\tanh$ and $\exp$ in stochastic domain. It is worth mentioning that both input and output of the Stanh function are in bipolar format, while the input and output of the Sexp function are in bipolar and unipolar formats respectively.

\begin{figure}[!t]
    \centering
    \subfigure[]{
        \includegraphics[scale = 1]{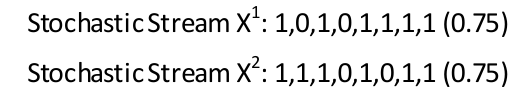}
        \label{75}
    }
    \subfigure[]{
        \includegraphics[scale = 1]{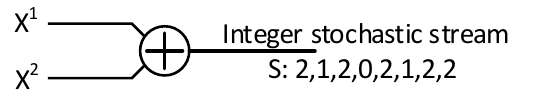}
        \label{NON}
    }
    \caption{(a) Stochastic representations of 0.75 and (b) Integer stochastic representation of 1.5}
    \label{nonbinary}
\end{figure}

\section{Proposed Integral Stochastic Computing}\label{sec2}
\subsection{Generation of Integer Stochastic Stream}\label{subsec4}
An integer stochastic stream is a sequence of integer numbers which are represented by either 2's complement or sign and magnitude. The average value of this stream is a real number $s \in$ $[0, m]$ for unipolar format and $s \in$ $[-m, m]$ for bipolar format, where $m \in \{1, 2, \dots \}$. In other words, the real value $s$ is the summation of two or more binary stochastic stream probabilities. For instance, 1.5 can be expressed as 0.75 + 0.75. Each of these probabilities can be represented by a conventional binary stochastic stream as shown in Fig.~\ref{75}. Therefore, the integer stochastic representation of 1.5 can be readily achieved as a summation of generated binary stochastic streams as illustrated in Fig.~\ref{NON}. In general, the integer stochastic stream $S$ representing the real value $s$ is a sequence with elements $S_i, i = \{1, 2, \dots, N\}$:
\begin{equation}
S_i = \sum_{j=1}^{m} X_i^j,
\label{start}
\end{equation}
where $X_i^j$ denotes each element of a binary stochastic sequence representing a real value $x^j$. The expected value of the integer stochastic stream is then given by
\begin{equation}
s = \E[S_i] = \sum_{j=1}^{m}x^j.
\label{eq1}
\end{equation}

\begin{figure}[!t]
    \centering
    \subfigure[]{
        \includegraphics[scale = 1]{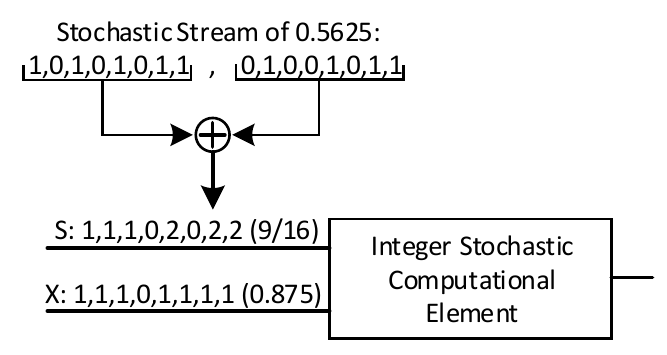}
        \label{latency}
    }
    \subfigure[]{
        \includegraphics[scale = 1]{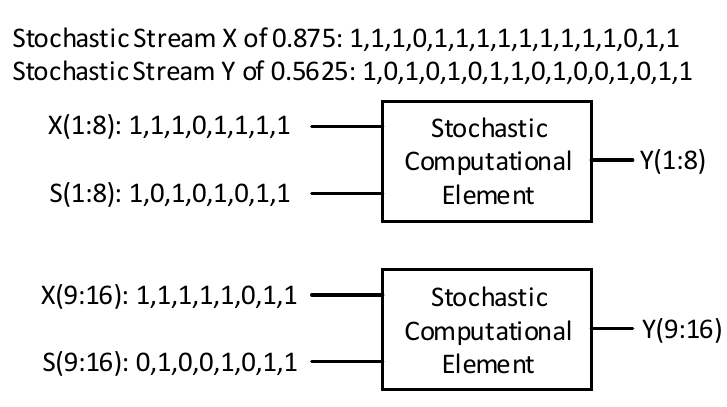}
        \label{parallel}
    }
    \caption{(a) Increasing the range value $m$ of the integer stochastic stream reduces computations latency. (b) Parallelized stochastic computation by factor of two.}
\end{figure}

We can also generate integer stochastic streams in the bipolar format. In that case, the elements $S_i$ of the stream are given by:
\begin{equation}
S_i = 2 \times \sum_{j=1}^{m} X_i^j - m,
\end{equation}
and the value represented by the stream is
\begin{equation}
s = \E[S_i] = 2 \times \sum_{j=1}^{m} \E[X_i^j] - m = 2\times\sum_{j=1}^{m}x^j - m.
\end{equation}



Any real number can be approximated by using an integer stochastic stream without prior scaling, as opposed to a conventional stochastic stream which is restricted only to the [-1, 1] interval. In integral SC, computation on two streams with different effective length is also possible while conventional SC fails to provide this property. For instance, representation of 0.875 and 0.5625 require effective bit-stream lengths of 8 and 16, respectively, using conventional SC. Therefore, effective bit-stream lengths of 16 is used to generate the conventional stochastic bit-stream of these two numbers for operations. However, the second number which requires higher effective length, i.e., 0.5625 in this example, can be generated by using the proposed integral SC with $m = 2$ as shown in Fig.~\ref{latency}. In this case, the bit-stream length of 8 is used for both numbers and operations can be performed by using lower lengths w.r.t. conventional SC. This technique potentially reduces the latency brought by stochastic computations, making integral SC suitable for throughput-intensive applications. It is worth mentioning that the integral SC is different from the conventional parallelized SC \cite{par}. For the sake of clarity, the aforementioned example is illustrated in Fig.~\ref{parallel} by using the conventional parallelized SC by factor of two. This is due to the fact that if several copies of a binary SC system are instantiated, the inputs still need to have the same effective length.\par
In summary, a real number $s \in [0, m]$ is first divided into the summation of multiple numbers which are in $[0, 1]$ interval. Then, the integer stochastic stream of this number is generated by using column-wise addition (see equations (\ref{start})-(\ref{eq1})). The bipolar format of the integer stochastic stream is generated in a similar way. Note that the binary to integer stochastic convertor is hereafter referred to as B2IS and it is composed of $m$ B2S convertors followed by and adder as shown in Fig.~\ref{nonbinary}.

\begin{figure}[!t]
    \centering
    \subfigure[]{
        \includegraphics[scale = 0.8]{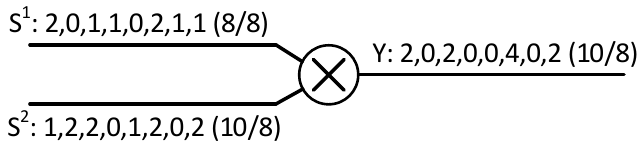}
        \label{multiplier}
    }
    \subfigure[]{
        \includegraphics[scale = 0.8]{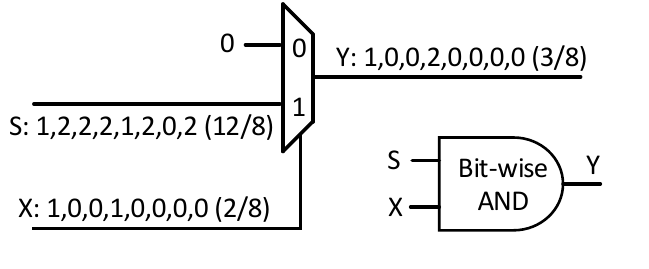}
        \label{multiplier1}
    }
    \caption{(a) Integer stochastic multiplier with $m = 2$ (b) Multiplication of integer stochastic stream with binary stochastic bit-stream using \textsc{and} gate or MUX}
    \label{nonmultiplier}
\end{figure}\par

\subsection{Implicit Scaling of Integer Stochastic Stream}\label{subsec_trick}
The integer stochastic representation of a real number $s \in [0,1]$ can also be generated by using an implicit scaling factor. In this method, the expected value of the individual binary streams is chosen as $x^j = s$, and the value $s$ represented by the integer stream is given by
\begin{equation}
s = \dfrac{\E[S_i]}{m}.
\end{equation}
This method avoids the need to divide $s$ by $m$ to obtain $x^j$, and can be easily taken into account in subsequent computations.
For instance, a real number $9/16$ can be represented using an integer stream length of $8$ with $m=2$. We can set $x^j=9/16$ (with an implicit scaling factor of $1/2$) and generate two binary sequences of length $8$. These sequences are then added together to form the integer sequence $S$. We obtain $\E[S_i] = 9/8$, which corresponds to $s=9/16$ because of the implicit scaling factor of $1/2$ (see Fig.~\ref{latency}).



\subsection{Multiplication In Integral SC}\label{subsec5}
The main advantage of SC compared to its binary radix format 
is the low complexity implementation of mathematical operations.
It is shown that multiplication can be implemented by using \textsc{and} or \textsc{xnor} gates depending on the coding format. However, integer stochastic multipliers make use of binary radix multipliers (see Fig.~\ref{multiplier}). The multiplication of two real numbers $s^1 \in [0, m]$ and $s^2 \in [0, m']$ with integer stochastic streams $S^1$ and $S^2$ in unipolar format is performed as follows:
\begin{equation}
y = s^1 \times s^2 = \E[S^1_i \times S^2_i] = \E[S^1_i] \times \E[S^2_i],
\end{equation} if $S^1_i$ and $S^2_i$ are independent.\par

The above equation holds true for integer stochastic multiplication in bipolar format as well. The implementation cost of this multiplier strongly depends on $m$ and $m'$. Considering one of these two values to be equal to "1", the multiplication can be implemented using bit-wise \textsc{and} gate or a MUX as depicted in Fig.~\ref{multiplier1}. 
The range of $y$ is $[0, m\times m']$ in the unipolar case, and $[-m\times m', m\times m']$ in the bipolar case.


\begin{algorithm}[!t]
\label{alg1}
 \KwData{Stochastic stream $X_i \in \{0, 1\}$ where $i \in \{1, 2, ..., N\}$}
 \KwResult{$Y_i$ }
 $Counter$ $\gets$ $Initial~value$\;
 \For{$i \gets 1$ : $N$}{
  $Counter$ $\gets$ $Counter$ + 2$X_i$ - 1\;
  \If{$Counter$ > $n$-1}{
  $Counter$ $\gets$ $n$-1\;}
  \If{$Counter$ < 0}{
   $Counter$ $\gets$ 0\;}
   \eIf{$Counter$ > $offset$}{
   $Y_i$ $\gets$ 1\;}{
   $Y_i$ $\gets$ 0\;}
   }
 \caption{Pseudo code of the conventional algorithm for FSM-based functions}
\end{algorithm}\par

\begin{algorithm}[!t]
\label{alg2}
 \KwData{Integer value $S_i \in \{-m,..., m\}$ where $i \in \{1, 2, ..., N\}$}
 \KwResult{$Y_i$ }
 $Counter$ $\gets$ $Initial~value$\;
 \For{$i \gets 1$ : $N$}{
  $Counter$ $\gets$ $Counter$ + $S_i$\;
  \If{$Counter$ > $n\times m$-1}{
  $Counter$ $\gets$ $n\times m$-1\;}
  \If{$Counter$ < 0}{
   $Counter$ $\gets$ 0\;}
   \eIf{$Counter$ > $offset$}{
   $Y_i$ $\gets$ 1\;}{
   $Y_i$ $\gets$ 0\;}
   }
 \caption{Pseudo code of the proposed algorithm for integer stochastic FSM-based functions}
\end{algorithm}\par

\subsection{Addition In Integral SC}\label{subsec6}
Conventional SC suffers from precision loss incurred by using scaled adder, making SC inappropriate for applications which require many additions. On the other hand, integral SC uses binary radix adders to perform additions in this domain, preserving all information. Using (\ref{eq1}), addition in unipolar format is performed as follows:

\begin{equation}
\label{eq:addmod}
y = s^1 + s^2 = \E[s^1 + s^2] = \E[S^1_i] + \E[S^2_i],
\end{equation} since the expected value operator is linear.\par

Equation \ref{eq:addmod} remains valid also in the bipolar case, while the range of $y$ is $[0, m + m']$ and $[-(m + m'), m + m']$ for unipolar and bipolar formats respectively.
This adder provides some advantages similar to APC. First of all, due to the fact that it retains all information provided as inputs, it reduces the variance of the sum. Secondly, it potentially reduces the bit-stream length required for computations compared to conventional SC \cite{ref5}. Moreover, the output of this adder is still an integer stochastic stream, which can be used by subsequent stochastic computational units, as opposed to APC.

\begin{figure*}[!t]
    \centering
    \subfigure[]{
        \includegraphics[scale = 0.31]{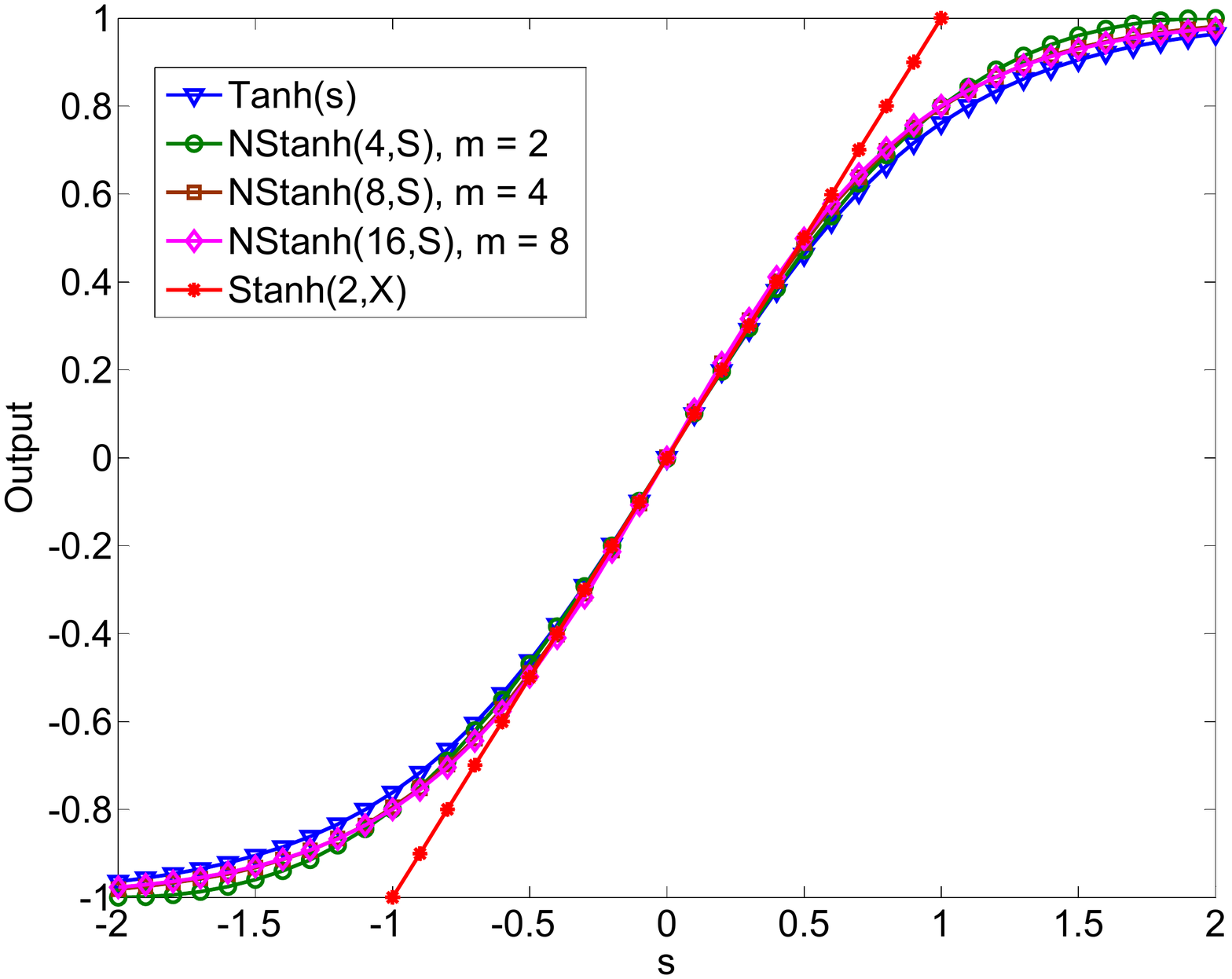}
        \label{MATLAB1}
    }
    \subfigure[]{
        \includegraphics[scale = 0.31]{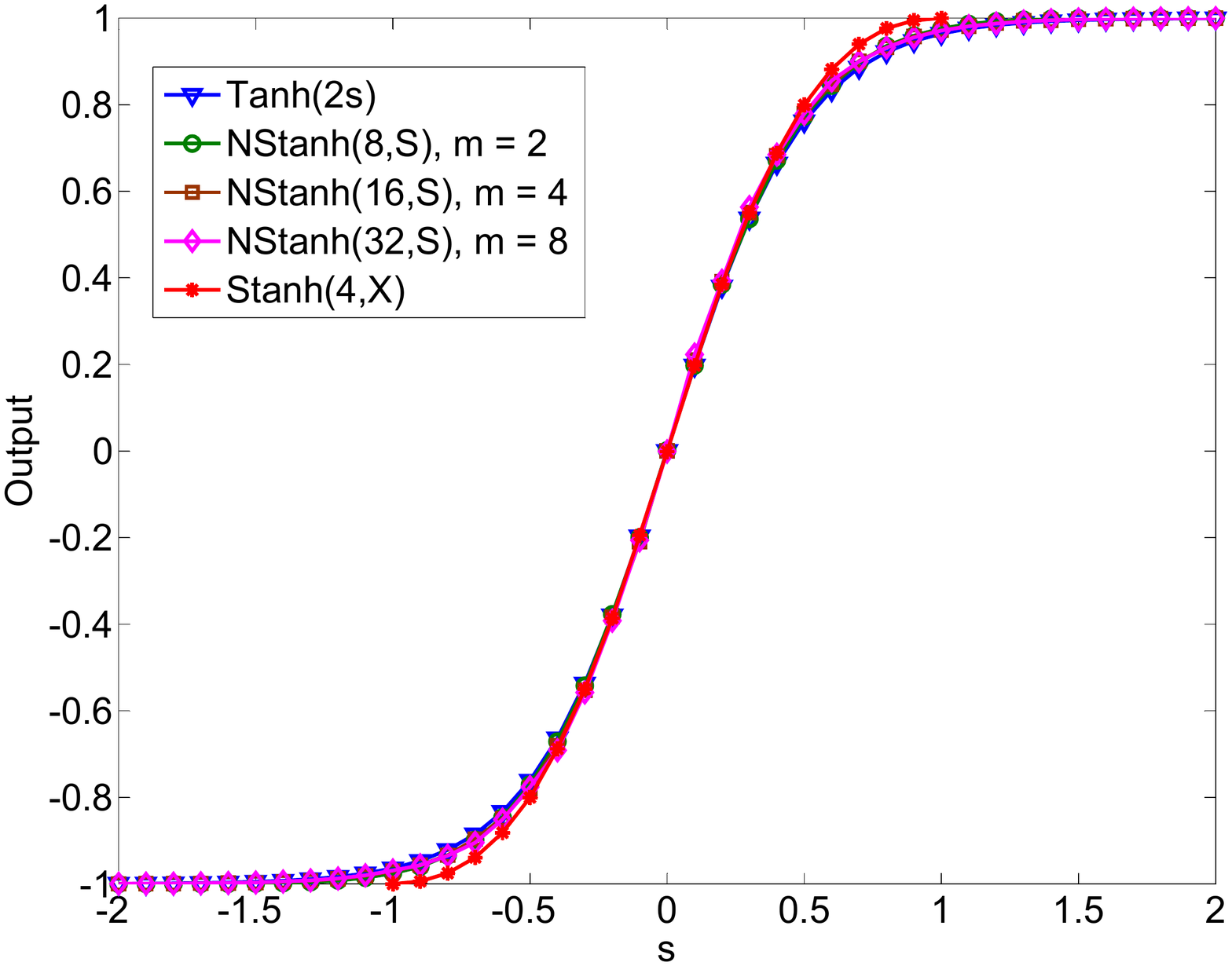}
        \label{MATLAB2}
    }
    \caption{(a) Integer stochastic implementation of $\tanh(s)$ and (b) Integer stochastic implementation of $\tanh(2s)$}
    \label{mattanh}
\end{figure*}

\begin{figure*}[!t]
    \centering
    \subfigure[]{
        \includegraphics[scale = 0.31]{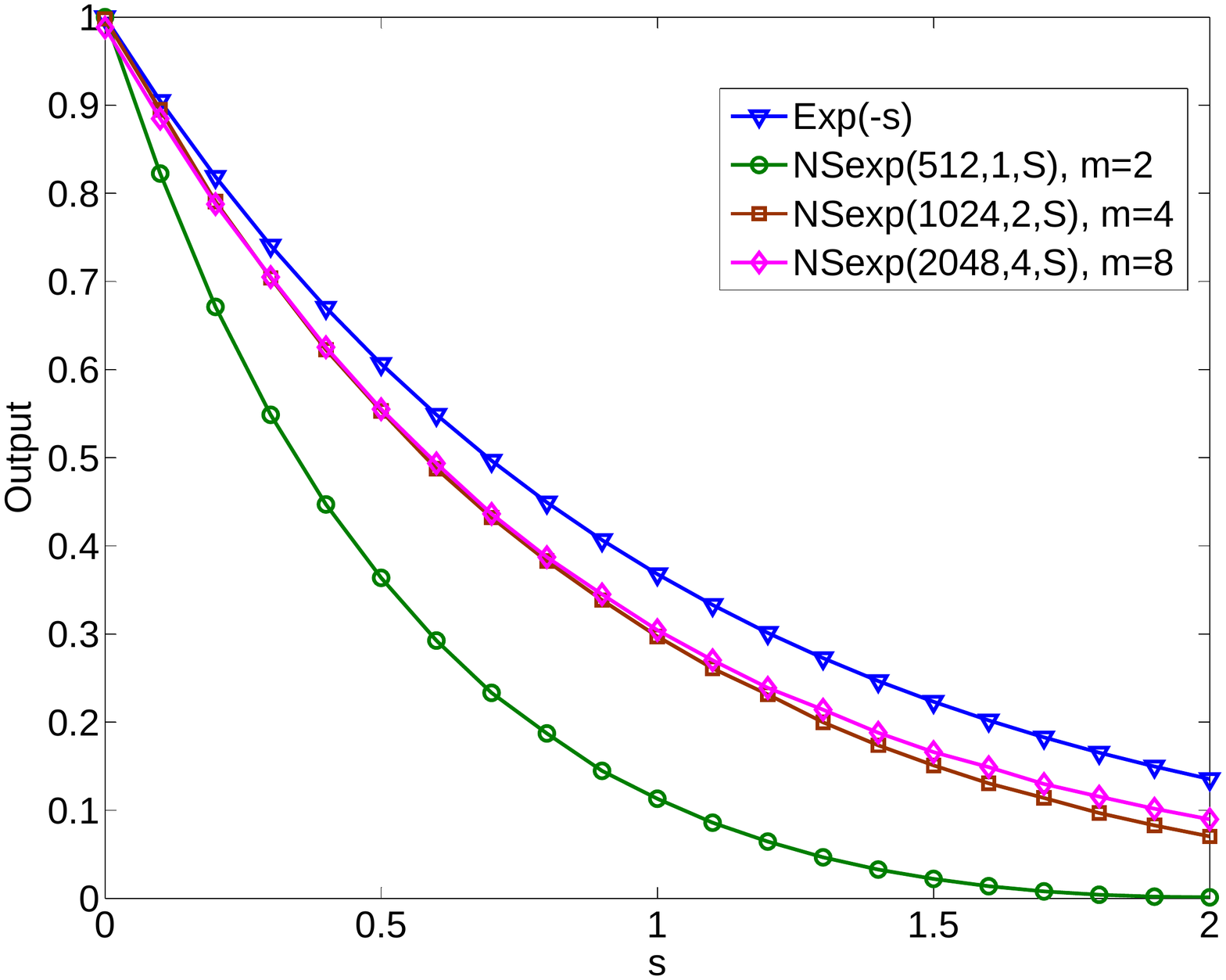}
        \label{MATLAB3}
    }
    \subfigure[]{
        \includegraphics[scale = 0.31]{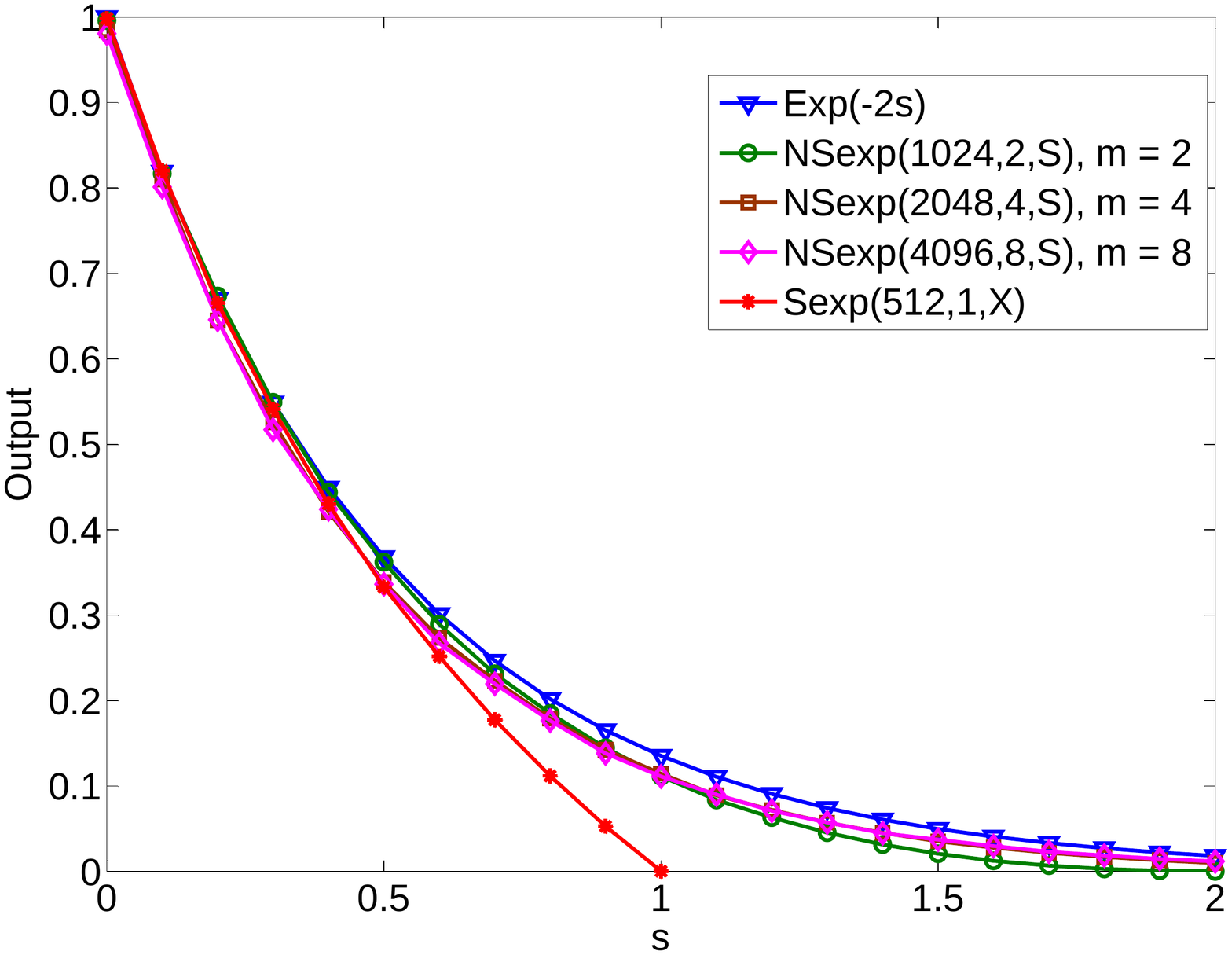}
        \label{MATLAB4}
    }
    \caption{(a) Integer stochastic implementation of $\exp(-s)$ and (b) Integer stochastic implementation of $\exp(-2s)$}
    \label{matexp}
\end{figure*}

\subsection{FSM-Based Functions In Integral SC}\label{subsec7}
The inputs of stochastic FSM-based $\tanh$ and exponentiation functions are restricted to real values in the [-1, 1] interval.
Therefore, a desired $\tanh$ or exponentiation function can be achieved by scaling down the inputs and adjusting the term $n$ in (\ref{tanh}) and (\ref{exp}), which potentially increases bit-stream length and results in long latency. The transition between each state of FSM is performed according to the input value in bipolar format, which is either 1 or 0. This state transition can be formulated as shown in Algorithm \ref{alg1} in conventional SC. According to the Algorithm \ref{alg1}, the input value in bipolar format is first converted to either 1 or -1 as an input of either 1 or 0, respectively. Then, the counter of FSM is added with the new encoded values which are similar to the values in an integral stochastic stream with $m = 1$. Therefore, the values of the conventional stochastic stream can be viewed as hard values of an integral stochastic stream. The FSM-based functions in integral SC can be achieved by extending the conventional FSM-based functions to support soft values in integral SC, which is explained below.\par

The integer stochastic $\tanh$ and exponentiation functions are proposed by generalizing Alg. \ref{alg1}. In integral SC, each element of a stochastic stream is represented using 2's complement or sign-magnitude representations in $\{-m, \dots, m\}$ for bipolar format. A state counter is increased or decreased according to the integer input value $S_i \in \{-m, \dots m\}$ where $i \in \{1, 2, ... ,N\}$. Therefore, the state counter is incremented or decremented by up to $m$ in each clock cycle, as opposed to conventional FSM-based functions which are restricted to one-step transitions. The algorithm for integer FSM-based functions is proposed as shown in Algorithm \ref{alg2}.


The output of the proposed integer FSM-based functions in integral SC domain and its encoding format are similar to the conventional FSM-based functions. For instance, the output of the integer $\tanh$ function is in bipolar format while the output of integer exponentiation function is in unipolar format. Moreover, the integer FSM-based functions require $m$ times more states compared to its conventional counterpart. Therefore, the approximate transfer function of integer $\tanh$ and exponentiation functions, which are referred to as NStanh and NSexp, respectively, are:
\begin{equation}
\tanh\left(\dfrac{ns}{2}\right) \approx \E[\text{NStanh}\left(m\times n,S\right)],
\label{tanh1}
\end{equation}
\begin{equation}
\exp\left(-2Gs\right) \approx \E[\text{NSexp}\left(m \times n,m \times G,S\right)] : s > 0.
\label{exp1}
\end{equation}\par

\begin{table*}[!t]
\renewcommand{\arraystretch}{1.3}
\renewcommand{\thefootnote}{\alph{footnote}}
\caption{Hardware Complexity of the Proposed FSM-Based Functions @ 400 MHz in a 65~nm CMOS Technology} 
\centering 
\begin{tabular}{c c c |c c| c c| c c} 
\hline\hline 
\multicolumn{1}{l|}{$m$ (Stream Length)} & \multicolumn{2}{c|}{1 (1024)}& \multicolumn{2}{c|}{2 (512)}& \multicolumn{2}{c|}{4 (256)}& \multicolumn{2}{c}{8 (128)}\\
\hline

\multicolumn{1}{l|}{} & Area ($\mu$m$^2$)& Power ($\mu$W)& Area ($\mu$m$^2$)& Power ($\mu$W)& Area ($\mu$m$^2$)& Power ($\mu$W)& Area ($\mu$m$^2$)& Power ($\mu$W)\\
\hline
\multicolumn{1}{l|}{$\tanh(s)$} & 24 &3.5 &74 &9.8 &117 &18.3 &150 &24.9\\
\hline
\multicolumn{1}{l|}{$\tanh(2s)$} & 63 &9.4 &107 &17.2 &141 &22.6 &182 & 31.2\\
\hline
\multicolumn{1}{l|}{$\exp(-s)$} & -- & --&424&62.1&474& 72.5& 480& 80\\
\hline
\multicolumn{1}{l|}{$\exp(-2s)$} & 424 & 57.1& 440& 65.1& 491& 74.5& 532& 93.1\\

\hline
\hline
\end{tabular}
\label{fsmtable} 
\end{table*}

In order to show the validity of the proposed algorithm, Monte-Carlo simulation is used. Fig.~\ref{mattanh} illustrates two examples of the proposed NStanh function compared to its corresponding Stanh and $\tanh$ functions for different values of $m$. Simulation results show that NStanh is more accurate than Stanh for $m > 1$ and that the accuracy improves as the value of $m$ increases. Moreover, NStanh is able to approximate $\tanh$ for input values outside of the [-1, 1] range with negligible performance loss, while Stanh does not work. 
The proposed NStanh function can also approximate $\tanh$ functions with fractional scaling factor, e.g. $\tanh \left(3/2x\right) \approx \text{NStanh}\left(3\times m,S\right)$, as long as the value $m$ is even, to make sure that the number of states is even. The aforementioned statements also hold true for NSexp, unlike with Sexp, as shown in Fig.~\ref{matexp}. The proposed FSM-based functions in integral SC also result in better approximation as the value of $n$ increases, similar to conventional stochastic FSM-based functions. The hardware complexity of the proposed FSM-based functions in a 65~nm CMOS technology is also summarized in Table~\ref{fsmtable}. The implementation results show that the proposed FSM-based functions consume roughly 7 times more power at most while having 8 times less latency, which results in a lower energy consumption, compared to the conventional FSM-based functions (i.e., FSM-based functions with $m = 1$). Note that the stream length of FSM-based functions denotes the latency.\par

\section{Integer Stochastic Implementation of DBN}\label{sec3}
\subsection{A Review on the DBN Algorithm}\label{subsec10}
DBNs are the hierarchical graphical models obtained by stacking RBMs on top of each other and training them in a greedy unsupervised manner \cite{ref1,ref2}. DBNs take low-level inputs and construct higher-level abstractions through the composition of layers. Both the number of layers and the number of inputs in each layer can be adjusted. Increasing the number of layers and their size tends to improve the performance of the network.\par

In this paper, we exploit a DBN constructed using two layers of RBM, which are also called hidden layers, followed by a classification layer at the end for handwritten digit recognition.. As a benchmark, we use the Mixed National Institute of Standards and Technology (MNIST) data set \cite{ref7}. This data set provides thousands of 28$\times$28 pixel images for both training and testing procedures. Each pixel is represented by an integer number between 0 to 255, requiring 8 bits for digital representation. As mentioned in Section \ref{introduction}, the training procedure can be performed on remote servers in the cloud. Therefore, the extracted weights are stored in a memory for the hardware inference engine 
to classify the input images in real-time.\par

\begin{figure}[!t]
\centering
\includegraphics[scale = 1]{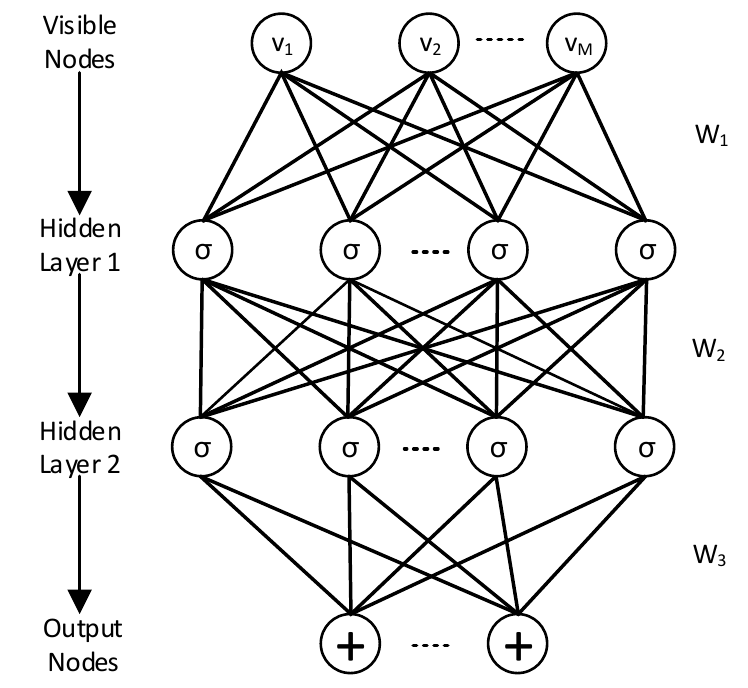}
\caption{The high-level architecture of 2-layer DBN.}
\label{DBN1}
\end{figure}

Fig.~\ref{DBN1} shows the DBN used for handwritten digits classification in this paper. Inputs of DBN and outputs of a hidden layer are hereafter referred to as visible nodes and hidden nodes, respectively. Each hidden node is also called neuron. The hierarchical computations of each neuron are performed as follows:
\begin{equation}
z_j = \sum_{i=1}^{M}W_{ij}v_i + b_j,
\label{matrix}
\end{equation}
\begin{equation}
h_j = \dfrac{1}{1+\exp(-z_j)} = \sigma(z_j),
\label{sigmoid}
\end{equation}
where $M$ denotes the number of visible nodes, $v_j$ the value of visible nodes, $W_{ij}$ the extracted weights, $b_j$ the bias term, $z_j$ intermediate value, $h_j$ the output value of each hidden node and $j$ an index to each hidden node. The nonlinearity function used in DBN , i.e., equation (\ref{sigmoid}), is called a sigmoid function. The classification layer does not require a sigmoid function as it is only used for quantization. In other words, the maximum value of the output denotes the recognized label.\par

\subsection{The Proposed Stochastic Architecture of a DBN}\label{subsec11}
VLSI implementations of a DBN network in binary form are computationally expensive since they require many matrix multiplications. Moreover, there is no straightforward way to implement the sigmoid function in hardware. Therefore, this unit is normally implemented by LUTs, which requires additional memory in addition to the memory used for storing weights. Considering 10 bits for weights, 78400 10b$\times$8b-multipliers are required to do the matrix multiplications of the first hidden layer for a parallel implementation of a network with configuration of 784-100-200-10, meaning 784 visible nodes, 100 first-layer hidden nodes, 200 second-layer hidden nodes and 10 output nodes. Note that the parallel implementation of such a networks results in huge silicon area in part due to its routing congestion caused by the layer interconnection.\par

\begin{figure}[!t]
\centering
\includegraphics[scale = 0.8]{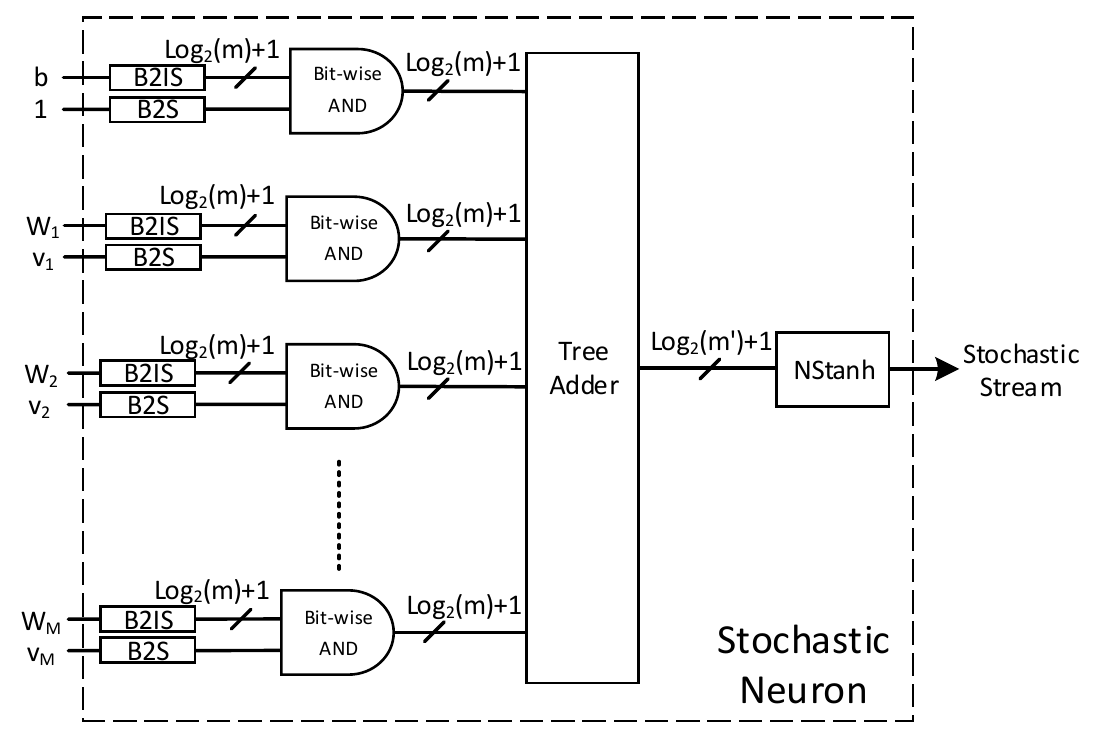}
\caption{The proposed integer stochastic neuron. The B2IS and B2S denote binary to integer stochastic and binary to stochastic converters, respectively.}
\label{stocharc}
\end{figure}

Stochastic implementation of DBN is a promising approach to perform the mentioned complex arithmetic operations using simple and low-cost elements. In order to find the output value of the first hidden node, 784 multiplications are required, which can be easily performed by using \textsc{and} gates in unipolar format. Then, addition of multipliers output should be performed by using a scaled adder or an \textsc{or} gate. Using a scaled adder to sum 784 numbers requires an extremely long bit-stream due to the fact that the output result of this adder is scaled down by 784 times, a very small number to be represented by short stream length.
In \cite{ref8}, an \textsc{or} gate is used as an adder to perform this computation while the inputs first are scaled down to make the term "$A\cdot B$" close to 0 in (\ref{add1}), which potentially increases the required stream length for computations. An APC is also proposed in \cite{ref5} to realize the matrix operations. Despite its good performance on additions, it is not a suitable approach for a stochastic DBN, since it converts the results to a binary form \cite{ref5}.\par

We have shown in Section~\ref{subsec4} that the integer stochastic stream can be generated by adding conventional stochastic streams. Considering that the multiplications of the first layer of a DBN are performed in conventional stochastic domain, the nature of the algorithm is to add the multiplication results together. Exploiting a binary tree adder, the addition result remains in integer-stochastic form without any precision loss. The sigmoid function can also be implemented in the integer stochastic domain.\par
It is well-known that the sigmoid function can be computed using the $\tanh$ function as follows:
\begin{equation}
\sigma(x) = \dfrac{1+\tanh\left(\dfrac{x}{2}\right)}{2}.
\label{sigtan}
\end{equation}
The $\tanh$ function can also be implemented by NStanh function (see (\ref{tanh1})) in integer stochastic domain. The output of NStanh is in bipolar format in conventional stochastic domain. Therefore, considering its output in unipolar format according to (\ref{sigtan}) and (2), the output of NStanh is equivalent to the sigmoid function in stochastic domain.\par

\begin{figure}[!t]
\centering
\includegraphics[scale = 0.18]{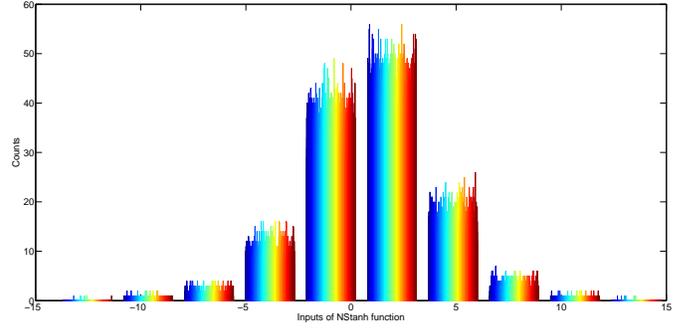}
\caption{Histogram of integer values as inputs of NStanh function at the first layer of a 784-100-200-10 DBN.}
\label{hist}
\end{figure}\par

Fig.~\ref{stocharc} shows the proposed integer stochastic architecture of a single neuron. The input signal stream is generated by using conventional stochastic domain: however, the weights are represented by 2's complement format in integer stochastic domain with range of $m$, which requires $\log2\left(m\right)+1$ bits for representation. The multiplications are performed bit-wise by \textsc{and} gates since pixels and weights are represented by binary stochastic streams and integral stochastic streams, respectively. A tree adder and an NStanh unit are used to perform the additions and nonlinearity function, respectively. The output of the integer stochastic sigmoid function is represented by a single wire in unipolar format. Therefore, the input and output formats are the same. Integer stochastic architecture of DBN is formed by stacking the proposed single neuron architecture. The input images require a minimum bit-stream length of 256, but since the weights lie in the $[-4, 4]$ interval they require a minimum bit-stream length of 1024 in conventional stochastic domain. Therefore, the latency of the proposed integer-stochastic implementation of the DBN is equal to 1024 for $m = 1$.\par

\begin{table}[!t]
\renewcommand{\arraystretch}{1.3}
\renewcommand{\thefootnote}{\alph{footnote}}
\caption{The Misclassification Error of the Proposed Architectures for Different Network Sizes and Stream Lengths} 
\centering 
\begin{tabular}{c c c c c} 
\hline\hline 
\multicolumn{1}{l|}{} & \multicolumn{4}{c}{Misclassification Error (\%)}\\
\hline
\multicolumn{1}{l|}{} & \multicolumn{1}{c|}{\cite{ref23}}& \multicolumn{3}{c}{Proposed}\\
\hline
\multicolumn{1}{l|}{Code Type} & \multicolumn{1}{c|}{Floating Point} & \multicolumn{3}{c}{Integeral SC}\\
\hline
\multicolumn{1}{l|}{$m$} & \multicolumn{1}{c|}{--} & 1 & 2 & 4\\
\hline
\multicolumn{1}{l|}{Stream Length} & \multicolumn{1}{c|}{--} & 1024 & 512 & 256\\
\hline
\multicolumn{1}{l|}{784-100-200-10} & \multicolumn{1}{c|}{2.29} & 2.40 & 2.46 & 2.33\\
\hline
\multicolumn{1}{l|}{784-300-600-10} & \multicolumn{1}{c|}{1.82} & 2.01 & 1.89 & 1.90\\
\hline
\hline
\end{tabular}
\label{table1} 
\end{table}

\begin{table*}[!t]
\renewcommand{\arraystretch}{1.3}
\renewcommand{\thefootnote}{\alph{footnote}}
\caption{Implementation Results of the Proposed Architecture on FPGA Virtex-7} 
\centering 
\begin{tabular}{c c c c c c c} 
\hline\hline 
\multicolumn{1}{c}{} & Network Size & Stream Length & Misclassification Error & Area (\# of LUTs) & Latency ($\mu$s) & Throughput (Mbps)\\  
\hline
\multicolumn{1}{l}{}  & 784-100-200-10 & 256 & 2.33\% & 1,013,002 & 1.705 & 3822 \\ 

\multicolumn{1}{l}{Proposed} & 784-100-200-10 & 512 & 2.46\% & 682,352 & 3.412 & 1874 \\

\multicolumn{1}{l}{}   & 784-100-200-10 & 1024 &  2.40\% & 437,461& 6.503 & 974\\
\hline
\multicolumn{1}{l}{}   & 784-100-200-10 & 1024 &  5.72\% & 144,450 & 8.561& NA\\

\multicolumn{1}{l}{\cite{ref8}}   & 784-300-600-10 & 1024 &  2.92\% & 603,750 & 9.797& NA\\

\multicolumn{1}{l}{}   & 784-500-1000-10 & 1024 &  2.32\% & 1,292,310 & 10.77& NA\\
\hline
\hline
\end{tabular}
\label{table2}
\end{table*}

The input range of the NStanh function, i.e. the value of $m'$ in Fig.~\ref{stocharc}, is selected through simulation. The histogram of the adder outputs identifies this range by taking a window which covers 95\% of data. For instance, Fig.~\ref{hist} shows the histogram of integer values as inputs of NStanh function at the first layer of a 784-100-200-10 DBN. This diagram is generated based on the non-correlated stochastic inputs and the selected range for this network is 6, i.e., the value of $m'$ in Fig.~\ref{stocharc}. This range strongly depends on the correlations among the stochastic inputs. The range would be a bigger number as the correlation increases.  For instance, summation of two correlated stochastic streams, $\{1, 1, 0, 0, 1, 0\}$ and
$\{1, 1, 0, 1, 0, 0\}$, representing real value of 0.5 results in integral stochastic stream of $\{2, 2, 0, 1, 1, 0\}$ and input range of 2 while summation of two uncorrelated stochastic streams, $\{0, 0, 1, 0, 1, 1\}$ and $\{1, 1, 0, 1, 0, 0\}$, representing real value of 0.5 results in integral stochastic stream of $\{1, 1, 1, 1, 1, 1\}$ and input range of 1. Correlation among the inputs is introduced when the same LFSR units are shared among several inputs, in order to reduce hardware area. In this paper, the set of LFSR units that are used for one neuron are shared for all the other neurons. More precisely, 785 11-bit LFSRs with different seeds are used in total to generated all inputs and weights of the proposed DBN architectures and guarantee
non-correlated stochastic streams.\par

\section{Implementation and Simulation Results}\label{sec4}
\subsection{Misclassification Error Rate Comparison}\label{subsec8}
The misclassification error rate of DBNs plays a crucial role in the performance of the system. In this part, the misclassification errors of the proposed integer stochastic architectures of DBNs with different configurations are summarized in Table~\ref{table1}. Simulation results have been obtained by using MATLAB on 10000 MNIST handwritten test digits \cite{ref7} for both floating point code and the proposed architecture using LFSRs as the stream generators. The method proposed in \cite{ref23} is used as our training core to extract the network weights. In fixed-point format, a precision of 10 bits is used to represent the weights. A stochastic stream of equivalent precision requires a length of 1024. The length of the stream can be reduced by increasing $m$. For example, using $m=2$ the length can be reduced to 512, and using $m=4$ it can be reduced to 256.
Because the input pixels only require 8 bits of precision, they can be represented using a binary ($m=1$) stochastic stream of length 256. Therefore, by using $m=1$ for the pixels and $m=4$ for the weights, it is possible to reduce the stream length to 256 while still using $\textsc{and}$ gates to implement multiplications. The simulation results show  the negligible performance loss of the proposed integer stochastic DBN for different sizes compared to their floating point versions. The reported misclassification errors for the proposed integral stochastic architecture were obtained using LFSR units as random number generators in MATLAB.

\subsection{FPGA Implementation}\label{subsec9}

\begin{table}[!t]
\renewcommand{\arraystretch}{1.3}
\renewcommand{\thefootnote}{\alph{footnote}}
\caption{ASIC Implementation Results For a 784-100-200-10 Network @ 400 MHz and 1V In a 65~nm CMOS Technology} 
\centering 
\scalebox{0.9}{
\begin{tabular}{c c c c | c} 
\hline\hline 
\multicolumn{1}{l|}{Implementation Type} & \multicolumn{3}{c|}{Integral SC}& Binary Radix\\
\hline
\multicolumn{1}{l|}{Stream Length} & 256 & 512 & 1024 & --\\
\hline
\multicolumn{1}{l|}{Misclassification error [\%]} & 2.33 & 2.46 & 2.40 & 2.3\\
\hline
\multicolumn{1}{l|}{Energy [$\mu$J]} & 2.96 & 3.3 & 3.35 & 0.380\\
\hline
\multicolumn{1}{l|}{Gate Count [M Gates (NAND2)]} & 4.2 & 2.2 & 1.1 & 23.6\\
\hline
\multicolumn{1}{l|}{Latency [ns]} & 650 & 1290 & 2570 & 30\\

\hline
\hline
\end{tabular}}
\label{energy} 
\end{table}\par

As mentioned previously, a fully- or semi-parallel VLSI implementation of DBN in binary form requires a lot of hardware resources. 
Therefore, many works target FPGAs \cite{ref17,ref18,ref19,ref20,ref21,ref22}, but none manage to fit a fully-parallel deep neural network architecture in a single FPGA board. Recently, a fully pipelined FPGA architecture of a factored RBM (fRBM) was proposed in \cite{ref24}, which could implement a single layer neural network consisting of 4096 nodes using virtualization technique, i.e., time multiplex sharing technique, on a Virtex-6 FPGA board. However, the largest fRBM neural network achievable without virtualization is on the order of 256 nodes.\par
In \cite{ref8}, a stochastic implementation of DBN on a FPGA board is presented for different network sizes, however, this architecture cannot achieve the same misclassification error rate as a software implementation. Table~\ref{table2} shows both the hardware implementation and performance results  of the proposed integer stochastic architecture of DBN for different network sizes on a Virtex7 xc7v2000t Xilinx FPGA. The implementation results show that the misclassification error of the proposed architectures for network size of 784-100-200-10 is the same as for the largest network presented in \cite{ref8}, i.e., the network size of 784-500-1000-10, while the area of the proposed designs are reduced by 66\%, 47\% and 21\% for $m$ = 1, $m$ = 2 and $m$ = 4. Moreover, the latency of the proposed architectures are also reduced by 40\%, 63\% and 84\% for $m$ = 1, $m$ = 2 and $m$ = 4. Therefore, as the value of $m$ increases, the latency of the integer stochastic hardware is reduced and becomes suitable for throughput-intensive applications. Note that the reported areas in Table~\ref{table2} include the costs of B2S and B2IS units.\par

\begin{table*}[!t]
\caption{ASIC Implementation Results For a 784-300-600-10 Network Based on Integral SC @ 400 MHz and 1V In a 65~nm CMOS Technology}
\centering
\scalebox{0.9}{
\begin{tabular}{c c c c c |c c c c |c c c c|c }
\hline\hline 
\multicolumn{1}{l|}{Implementation Type} & \multicolumn{12}{c|}{Integral SC}& \multicolumn{1}{c}{Binary Radix}\\
\hline
\multicolumn{1}{l|}{Network Configuration} & \multicolumn{12}{c|}{784-300-600-10}& \multicolumn{1}{c}{784-100-200-10}\\
\hline
\multicolumn{1}{l|}{Value of $m$} & \multicolumn{4}{c|}{1}& \multicolumn{4}{c|}{2} & \multicolumn{4}{c|}{4} & \multicolumn{1}{c}{--}\\
\hline
\multicolumn{1}{l|}{Stream Length} & 64 & 128 & 256 & 512 & 32 & 64 & 128 & 256 & \textbf{16} & 32 & 64 & 128 & -- \\
\hline
\multicolumn{1}{l|}{Misclassification error [\%]} & 2.49 & 2.24 & 2.22 & 2.07 & 2.42 & 2.30 & 2.24 & 1.96 & \textbf{2.27}& 2.22 & 2.07 & 1.95 & 2.3\\
\hline
\multicolumn{1}{l|}{Energy [$\mu$J]} & 0.740 & 1.436 & 2.802 & 5.640 & 0.505 & 1.029 & 1.997 & 3.933 & \textbf{0.299} & 0.640 & 1.28 & 2.53 & 0.380\\
\hline
\multicolumn{1}{l|}{Gate Count [M Gates (NAND2)]} & 5.4 & 5.6 & 5.6 & 5.6 & 9.2 & 9.7 & 10.2 & 10.2 & \textbf{15.6} & 16.9 & 17.8 & 18.9 & 23.6\\
\hline
\multicolumn{1}{l|}{Latency [ns]} & 170 & 330 & 650 &  1290 & 90 & 170 & 330 & 650 & \textbf{50} & 90 & 170 & 330 & 30\\

\hline
\hline
\end{tabular}}
\label{energy1}
\end{table*}

\subsection{ASIC Implementation}\label{subsec14}
Table~\ref{energy} shows the ASIC implementation results for a fixed-point implementation of the network size of 784-100-200-10. Despite the improvements that the proposed architectures provide over previously proposed stochastic implementations, the stochastic implementations still uses more energy than the fixed-point implementation in 65~nm CMOS, even if the power consumption and area of a stochastic neuron are smaller. A similar result was also obtained in \cite{ref26} for stochastic implementations of image processing circuits.\par
In order to improve the energy consumption of the proposed stochastic architectures, we select a bigger network size with better misclassification rate and reduce the stream length to achieve roughly the same misclassification error rate as the binary radix implementation in Table~\ref{energy}. The implementation results of a 784-300-600-10 neural network based on integral SC for different stream lengths and values of $m$ are summarized in Table~\ref{energy1}. The implementation results show that the integral stochastic architecture for value of $m = 4$ and stream length of 16 at misclassification error rate of 2.3\% consumes 21\% less energy as well as 34\% less area compared to the binary radix implementation.

\subsection{Quasi-Synchronous Implementations}\label{subsec15}

\begin{table}[!t]
\renewcommand{\arraystretch}{1.3}
\renewcommand{\thefootnote}{\alph{footnote}}
\caption{Deviations of Layer-1 and Layer-2 Neurons For a 784-300-600-10 Network} 
\centering 
\begin{tabular}{c c c c | c c c} 
\hline\hline 
\multicolumn{1}{l|}{} & \multicolumn{2}{c}{Deviation (\%)}\\
\hline
\multicolumn{1}{l|}{} & \multicolumn{1}{c|}{Layer-1 Neuron}& \multicolumn{1}{c}{Layer-2 Neuron}\\

\hline
\multicolumn{1}{l|}{0.7V} & 17.90 & 15.41\\
\hline
\multicolumn{1}{l|}{0.75V} & 8.57 & 3.95 \\
\hline
\multicolumn{1}{l|}{0.8V} & 0.011 & $\approx 0$\\
\hline
\hline
\end{tabular}
\label{dev} 
\end{table}\par
In order to further reduce the energy consumption of the system, we also consider a \emph{quasi-synchronous} implementation, in which the supply voltage of the circuit is reduced beyond the critical voltage by permitting some timing violations to occur. Timing violations introduce deviations in the computations, but because the stochastic architecture is fault-tolerant, we can obtain the same classification performance by slightly increasing the length of the streams. This yields further energy savings without any compromise on performance.\par

We characterize the effect of timing violations on the algorithm by studying small test circuits that can be simulated quickly, using the same approach as in \cite{francois}. In the proposed architecture, the same processing circuit can be replicated several times to form each layer, depending on the required degree of parallelism. Therefore, we characterize the effect of timing violations on these small processing circuits: each neuron processor (one for each layer) is synthesized in a 65~nm CMOS technology and deviations are measured at different voltages, from 0.7V to 1.0V in 0.05V increments, as shown in Table~\ref{dev}. Note that no deviations are observed when the supply voltage is larger than 0.8V. The output of first and second layers is binary, while the output of classification layer has 6 bits. Binary to stochastic converter units are also considered for each neuron and the weights are hard coded for the implementations.\par

\begin{table}[!t]
\renewcommand{\arraystretch}{1.3}
\renewcommand{\thefootnote}{\alph{footnote}}
\caption{ASIC Implementation Results For a 784-300-600-10 Network @ 400 MHz In a 65~nm CMOS Technology Under Faulty Conditions} 
\centering 
\scalebox{0.8}{
\begin{tabular}{c c c c} 
\hline\hline 
\multicolumn{1}{l|}{Implementation Type} & \multicolumn{3}{c}{Integral SC}\\
\hline
\multicolumn{1}{l|}{Supply Voltage (Layer-1--layer-2--layer-3)} & \multicolumn{1}{c}{0.8--0.7--0.8}& \multicolumn{1}{c}{0.75--0.75--0.8} & \multicolumn{1}{c}{0.8--0.8--0.8}\\
\hline
\multicolumn{1}{l|}{Stream Length} & 30 & 30 & 22\\
\hline
\multicolumn{1}{l|}{Misclassification error [\%]} & 2.29 & 2.28 & 2.30\\
\hline
\multicolumn{1}{l|}{Energy [$\mu$J] (improvement w.r.t. 1V)} & 0.283 (-5\%) & 0.286 (-4\%) & 0.256 (-14\%)\\
\hline
\multicolumn{1}{l|}{Gate Count [M Gates (NAND2)]} & 15.6 & 15.6 & 15.6\\
\hline
\multicolumn{1}{l|}{Latency [ns]} & 85 & 85 & 65\\
\hline
\hline
\end{tabular}}
\label{energy2} 
\end{table}

The deviation error of the layer-3 neuron for 0.7V and 0.75V results in a huge misclassification error. It is not beneficial to allow large deviations to occur in that layer since there are only 10 neurons in the third layer, and therefore we do not expect the supply voltage of layer-3 processing circuits to have a big impact on the overall energy consumption. Therefore, the layer-3 neurons supplied with 0.8V are used. Note that no deviations are observed when the supply voltage is 0.8V in the layer-3 neurons.\par 
The performance results for a 784-300-600-10 network and $m=4$ at different supply voltages are provided in Table~\ref{energy2}. The misclassification performance obtained by the quasi-synchronous system is very similar to the performance of the reliable system, despite the fact that the deviation rate is up to 9\% in layer-1 neurons and 16\% in layer-2 neurons. This results in up to a 14\% lower energy consumption without any compromise on performance. On the other hand, introducing bit-wise deviations at a rate of 1\% in the fixed-point system results in a 87\% misclassification rate. Note that the reported implementation results in this paper include costs of B2N and B2IS units.\par

Moreover, because a stochastic implementation is much more fault-tolerant than a fixed-point implementation, it can be preferable for future process technologies, and in particular for inherently unreliable ones such as nanoscale memristor devices. Note that memristor devices consume substantially less energy compared to CMOS and can be scaled to sizes below 10~nm \cite{ref25}. In \cite{ref25}, stochastic implementations were suggested as a promising approach for use in such devices.




\section{Conclusion}\label{sec5}
Integral SC makes the hardware implementation of precision-intensive applications feasible in the stochastic domain, and allows computations to be performed with streams of different lengths, which can improve the latency of the system. An efficient stochastic implementation of a deep belief network is proposed using integral SC. The simulation and implementation results show that the proposed design reduces the area occupation by $66\%$ and the latency by $84\%$ with respect to the state of the art. We also showed that the proposed design consumes 21\% less energy than its binary radix counterpart. Moreover, the proposed architectures can save up to 33\% energy consumption w.r.t. the binary radix implementation by using quasi-synchronous implementation without any compromise on performance.
\section*{Acknowledgement}
The authors would like to thank C. Condo for his helpful suggestions.
\bibliographystyle{IEEEtran}
\bibliography{Bibliography}

\end{document}